\documentclass[letterpaper, 10 pt, conference]{ieeeconf}  

\IEEEoverridecommandlockouts                              

\usepackage{amsmath, amsfonts, amssymb}
\usepackage{graphicx}
\usepackage{xcolor}
\usepackage{float}
\usepackage{cite}
\usepackage{bm}

\usepackage{algorithm}
\usepackage{algpseudocode}

\usepackage{booktabs}
\usepackage{multirow}

\usepackage{hyperref}

\newcommand{\mat}[1]{\bm{#1}}
\renewcommand{\vec}[1]{\bm{#1}}

\newcommand{\real}{\mathbb{R}}

\newcommand{\x}{\vec{x}} 
\newcommand{\f}{\mathbf{f}} 

\newcommand{\p}{\vec{p}}

\newcommand{\C}{\mat{C}} 
\newcommand{\B}{\mat{B}} 
\newcommand{\J}{\mat{J}} 

\renewcommand{\B}{\mat{B}}
\renewcommand{\C}{\mat{C}}

\newcommand{\btau}{\vec{\tau}}

\newcommand{\half}{\frac{1}{2}} 
\newcommand{\T}{^\top}          

\DeclareMathOperator*{\argmin}{argmin}

\DeclareMathOperator{\des}{des}

\graphicspath{{figures/}}

\algrenewcommand\algorithmicrequire{\textbf{Input:}}
\algrenewcommand\algorithmicensure{\textbf{Output:}}

\usepackage{mathtools}

\newif\ifanonymous
\anonymousfalse         

\renewcommand{\d}{\bm{d}} 
\newcommand{\LS}{\mathcal{L}} 
\newcommand{\DS}{\mathcal{D}} 

\title{\LARGE \bf
SurGE: Surrogate Gradient-guided Evolution for Co-design of \\
Legged Robots with Parallel Elasticity
}

\ifanonymous
\author{Anonymous Authors}
\else
\author{
Yulun Zhuang$^1$, Yue Qin$^1$, Justin Lu$^1$, Zelin Shen$^1$, Yichen Wang$^1$, Sicheng He$^2$ and Yanran Ding$^{1, *}$%
\thanks{$^1$Authors are with the Department of Robotics, University of Michigan, Ann Arbor, MI 48109, USA.
}%
\thanks{$^2$Sicheng He is with the Department of Mechanical and Aerospace Engineering, University of Tennessee, Knoxville, TN 37996, USA.
}%
\thanks{
$^*$This project was supported by the National Science Foundation under grant 2427036.
Corresponding author email: {\tt\small yanrand@umich.edu}
}%
}
\fi

\begin{document}

\maketitle
\thispagestyle{empty}
\pagestyle{empty}

\begin{abstract}

Co-design of legged robots with elastic elements is challenging due to the non-differentiability of contact dynamics and mechanism engagement.
This paper presents SurGE, a framework that computes surrogate gradients of the design objective through a differentiable pipeline consisting of a kinodynamic single-rigid-body (Kino-SRB) model and a design-aware control policy, and injects them into CMA-ES via mean shift with cosine-annealed step decay.
On a 4-DOF design space of a hopping robot with unidirectional parallel spring, SurGE achieves $6\times$ lower cross-seed standard deviation and $18\%$ tighter population concentration compared to vanilla CMA-ES, while matching or improving the best objective.
Hardware experiments on a 2D design subspace show that, starting from a hand-tuned initial design, SurGE reduces the design objective by $37.65\%$ on hardware, with the improvement trend identified in simulation transferring consistently to the physical system.
SurGE provides the potential to accelerate non-differentiable co-design problems in legged robots via surrogate model gradients.
\end{abstract}

\section{Introduction}

Legged robots have increasing potential in field applications such as planetary exploration and search-and-rescue scenarios.
Actuator energy efficiency remains a primary bottleneck for those deployments where the battery capacity limits mission duration.
Researchers have explored ways to improve actuation energy efficiency by incorporating elastic elements.
Among these methods, the parallel elastic actuator (PEA), springs mounted in parallel with actuators, has been shown to reduce the energy cost of periodic locomotion by passively storing and releasing energy during the gait cycle~\cite{yesilevskiy2015comparison, folkertsma2012parallel, liu2024dualslide, tanfener2024elastic}.

Unidirectional parallel springs (UPS) is a special category of PEA where they provide torque assistance in one direction while remaining disengaged in the other~\cite{Liu2015switch, BadriSprwitz2022bird}. 
Compared with the bidirectional PEA, UPS can provide assistant torque for stance phases while allowing leg extension without resistance during swing phases.
However, the energy benefit of a UPS is sensitive to its design, as well as how the control policy exploits the stored elastic energy.
A poorly chosen spring design or control policy can negate the expected efficiency gain, making careful parameter selection essential.
Sequential approaches, in which the spring is designed first and the controller tuned afterward, can lead to suboptimal solutions since the design parameters and the control policy are coupled in the evaluation of energy efficiency.

This sensitivity motivates a concurrent optimization of hardware design and control policy, commonly named the co-design problem~\cite{dinev2022versatile,belmonte2022meta, bjelonic2023learning, bravo2024engineering, fadini2021computational}.
As the co-design problem jointly optimizes control signals and design parameters, it is often formulated as a bilevel optimization, where the outer loop searches over the design space for design parameters that minimize an objective, while the inner loop optimizes a controller for each candidate design.

\begin{figure}[t]
    \centering
    \includegraphics[width=1.0\columnwidth]{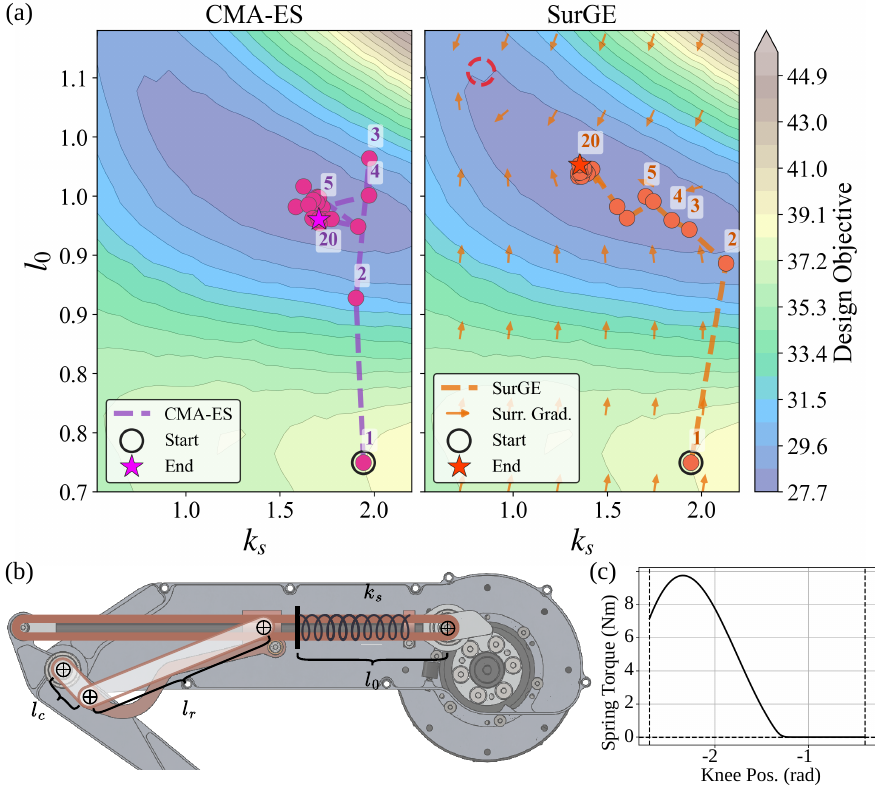}
    \caption{(a) Visualization of the population means over 20 optimization iterations of two example runs in the $(k_s, l_0)$ subspace. The objective landscape is shown in the background while the surrogate gradient vectors are overlaid for SurGE. 
    The trajectories of CMA-ES (left) and SurGE (right) are shown in purple and orange, respectively. Design parameter values are shown in scaled units. SurGE shows a tighter population concentration while being robust to a flat basin and biased gradient minimum (circled in red). (b) Linkage schematic of the UPS on the MUPS~v2 robot. The design space contains the spring stiffness $k_s$, the spring engagement length $l_0$, the length of rocker linkage $l_{r}$, and the length of crank linkage $l_{c}$. (c) The spring is modeled as a smoothed torque function w.r.t. knee joint positions.}
    \label{fig:cover_img_merged}
    \vspace{-0.5cm}
\end{figure}

Computational efficiency in co-design has long been a central concern, particularly for complex dynamical systems and high-dimensional design spaces. Since the inner loop must be optimized to converge for each candidate design parameter, it needs to be re-solved efficiently at every iteration. 
One common strategy is to utilize open-loop trajectories generated by motion planners to accelerate the inner-loop optimization~\cite{dinev2022versatile}.
However, such approaches often sacrifice solution accuracy due to the lack of high-fidelity dynamics simulation.
To retain the simulation fidelity, Bjelonic et al.~\cite{bjelonic2023learning} proposed a design-conditioned controller to approximate the solution of the inner optimization by training the mapping from design parameters to the corresponding near-optimal control policy. 
This approach effectively performs amortized optimization~\cite{amos2023tutorial}, replacing the per-design inner optimization with a single model trained to predict its solution. This allows the outer design search to proceed with the policy held fixed rather than retraining a controller for each candidate design.
Beyond the inner loop, another concern lies in the outer-loop objective, which is aggregated over an entire simulated trajectory. Such objectives are expensive to evaluate and often non-smooth due to ground contact and mechanism engagement, thereby limiting the overall computational efficiency in existing co-design frameworks.

Zero-order methods such as covariance matrix adaptation evolution strategy (CMA-ES)~\cite{hansen2006cma} are often used for the outer-loop design search for legged robots~\cite{bjelonic2023learning, belmonte2022meta, chadwick2020vitruvio}, as they treat the objective as a black box and make no smoothness assumptions~\cite{jordana2025introduction}.
These methods provide reliable solutions given a sufficient evaluation budget.
However, two practical limitations persist.
First, the sample complexity scales unfavorably with the design dimension, and each function evaluation requires a full simulation rollout, rendering the approach computationally prohibitive in high-dimensional design space. 
Second, because of its stochastic nature, the design returned by evolutionary search varies across random seeds, with different seeds often converging to different local optima~\cite{belmonte2022meta}. While exploring multiple optima can be valuable, committing to a single hardware design requires reproducible outcomes, which in practice forces practitioners to run many seeds and reconcile among them.

Gradient-based methods, such as gradient descent, could potentially accelerate convergence by leveraging the deterministic gradient information.
Nevertheless, computing accurate gradients of a legged locomotion task is non-trivial due to non-differentiable contact and mechanism engagement.
Trajectory optimization approaches which leverage an analytical dynamics model can provide smooth gradients, but typically require fixing contact schedules and employing simplified models that may not reflect the true dynamics~\cite{bravo2024engineering, chadwick2020vitruvio, he2026efficient, fadini2021computational}.
Differentiable simulators~\cite{freeman2021brax, qiao2021efficient, xu2022accelerated, schwarke2025learning} allow backpropagation through full-body dynamics, but require relaxed contact models to maintain differentiability, an approximation that can distort the ability of deploying a resulting controller onto a real robot~\cite{suh2022differentiable, metz2021gradients, song2024learning}.

Hybrid methods that combine zero-order and first-order methods have been proven to be useful. 
Even in the absence of accurate gradient information, surrogate gradients that are correlated with the true gradient can reduce variance and accelerate convergence.
Maheswaranathan et al.~\cite{maheswaranathan2019guided} demonstrated that augmenting the CMA-ES sampling distribution with an imperfect surrogate gradient subspace speeds up convergence.
Hansen~\cite{hansen2011injecting} proposed injecting externally generated solutions into CMA-ES, with the strongest variant being a mean shift that moves the distribution mean toward a target solution, biasing the entire population while preserving the diversity for the adaptation mechanism. 
Meanwhile, recent work on learning legged robot control policies via model-based reinforcement learning (RL)~\cite{song2024learning} proposes a method to compute the surrogate gradient for legged robot dynamics, enabling faster training convergence of deployable locomotion control policies. 
Motivated by these advances, combining surrogate gradients for legged robot dynamics with zero-order optimization methods in a hybrid framework presents a promising direction to improve the computational efficiency and scalability of legged robot co-design.

This paper presents SurGE (Surrogate Gradient-guided Evolution), a hybrid framework for the legged robot co-design problem that integrates the surrogate gradient with CMA-ES to achieve faster convergence and reduced variance during optimization. To address the non-differentiability of legged robot dynamics, a differentiable surrogate dynamics model is introduced to estimate the gradient information along the dynamics rollout, which aids CMA-EA in mean value shifting. The SurGE framework is evaluated on a planar hopping robot, Monoped with Unidirectional Parallel Spring v2 (MUPS~v2), demonstrating improved performance in both numerical simulations and hardware experiments.


The main contributions of this work are:
\begin{itemize}
    \item A co-design framework that enhances evolutionary strategy with surrogate gradient to accelerate design optimization for legged robots.
    \item Comprehensive benchmarks showing SurGE achieves lower cross-seed standard deviation, tighter population concentration, and faster convergence.
    \item Hardware deployment on the MUPS~v2 robot, confirming that SurGE refines a hand-tuned initial design and reduces the objective by $37.65\%$ on hardware.
\end{itemize}

\begin{figure*}
    \centering
\includegraphics[width=1.0\textwidth]{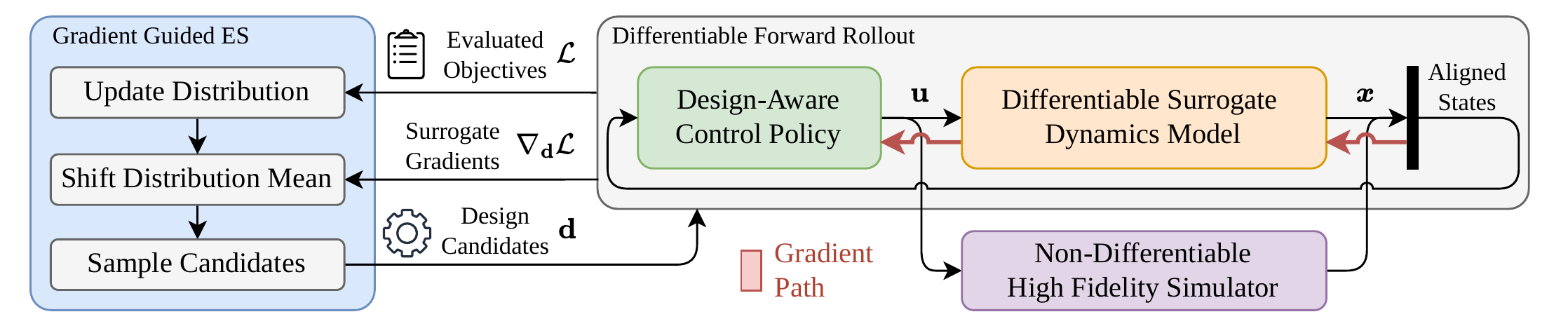}
    \caption{Overview of the SurGE framework. The surrogate gradient is computed through a differentiable pipeline consisting of a design-aware control policy (Sec.~\ref{sec:design_aware_policy}) and a kinodynamic single-rigid-body (Kino-SRB) model (Sec.~\ref{sec:gradient_computation}), and injected into CMA-ES via mean shift with cosine-annealed step decay (Sec.~\ref{sec:gradient_injection}). The true design objective is evaluated on a non-differentiable simulator, which also provides state alignment (Sec.~\ref{sec:state_alignment}) for the surrogate gradient computation.}
    \label{fig:differentiable_pipeline}
    \vspace{-0.3cm}
\end{figure*}

\section{Problem Statement}

The co-design problem can be described as a standard Optimal Control Problem (OCP) where the decision variables include states, control inputs, and design parameters. The discretized formulation for the co-design problem is:
\begin{align}
\min_{\bm{x},\bm{u},\bm{d}} \quad
&l_N(\bm{x}_N, \bm{d}) +   \sum_{k=0}^{N-1} l_k(\bm{x}_k,\bm{u}_k,\bm{d} )\\
\text{s.t. }\quad
& \bm{x}_{k+1} = \bm{f}(\bm{x}_k, \bm{u}_k, \bm{d})\\
& \bm{h}(\bm{x}_k, \bm{u}_k, \bm{d}) \le 0\\
& \bm{g}( \bm{x}_0, \bm{x}_N, \bm{d}) \le 0, 
\end{align}
where $\bm{x}_k \in \mathbb{R}^n$ and $\bm{u}_k \in \mathbb{R}^m (k=1, \ldots, N)$, represent the state and control vectors, respectively; $\bm{d}\in\mathbb{R}^{n_{\bm{d}}}$ denotes the design variables; $l_k$ represents the stage cost function; $l_N$ denotes the terminal cost function; $\bm{f}: \mathbb{R}^{n}\times \mathbb{R}^{m} \times \mathbb{R}^{n_{\bm{d}}}\rightarrow \mathbb{R}^{n}$ represents the dynamics of the system; $\bm{h}$ and $\bm{g}$ are the path constraints and boundary constraints, respectively; and $n, m, N$ and $n_{\bm{d}}$ represent the state, control, time, and design variables' dimensions, respectively.

To solve this OCP, a bilevel optimization strategy is commonly adopted. 
The outer level optimizes the design parameters with a fixed control policy, while the inner level optimizes the control policy for a given design. 
These two subproblems are solved alternately until convergence.
However, solving the inner problem to convergence for every candidate design parameter $\bm{d}$ is computationally expensive. 
To address this challenge, a design-aware optimal controller is introduced to approximate the solution of the inner optimization by taking the design parameters as input and maps them to the corresponding control policy~\cite{bjelonic2023learning, fadini2024making}.

Assuming that this design-aware policy, denoted by $\bm{\pi}^*:\mathbb{R}^{n}\times \mathbb{R}^{n_{\bm{d}}}\rightarrow \mathbb{R}^{m}$, which is a mapping from the current state and design variables to the control vector, achieves near-optimal performance across the whole design space $\DS$, the original co-design problem can be reformulated as a single-stage optimization problem, which is more tractable.

Within the formulation, the decision variables dimension will reduce to $n^{\bm{d}}$ as the control signal $\bm{u}_k$ could be determined given the system state $\bm{x}_k$ and design parameters $\bm{d}$.
As in prior co-design formulations~\cite{bravo2022robust}, an additional constraint representing the design-aware control policy is embedded into the OCP, i.e. $\bm{u}_k = \bm{\pi}^*\big(\bm{x}_k,\d\big)$.

Given this design-aware policy, the system becomes a closed-loop system where the dynamics could be simplified as a transition function that only depends on the last state and design parameter:
\begin{equation}
\begin{aligned}
    \bm{x}_{k+1} = \bm{f}\big(\bm{x}_k,\bm{\pi}^*(\bm{x}_k,\bm{d}),\bm{d}\big)
    \coloneq \bm{f}^{\prime}(\bm{x}_k,\bm{d}),
\end{aligned}
\end{equation}
where the $\bm{f}^{\prime}:\mathbb{R}^n \times \mathbb{R}^{n_{\bm{d}}} \rightarrow \mathbb{R}^n$ denotes this closed-loop system dynamics.
The cost function $l_k(\bm{x}_k,\bm{u}_k,\bm{d})$ could also be simplified as $l_k^{\prime}(\bm{x}_k,\bm{d})$ in a similar manner.
Within the formulation, the total cost $\LS$ could be expressed as:
\begin{equation}
    \LS = \sum_{k=0}^{N} l_k^{\prime}(\bm{x}_k,\bm{d} ).\label{eq:design_objective_high_level}
\end{equation}

According to the gradient analysis for iterated dynamical systems in \cite{metz2021gradients}, the gradient of total cost with respect to the design parameters $\frac{\partial \LS}{\partial \bm{d}} \in \mathbb{R}^{n_{\bm{d}}}$ is obtained by applying the chain rule through time:
\begin{equation}
    \frac{\partial \LS}{\partial \bm{d}} 
    =
     \sum_{k=0}^{N}\left(
        \frac{\partial l_k^{\prime}}{\partial \bm{d}} 
        +
        \sum_{i=1}^k \frac{\partial l^{\prime}_k}{\partial \x_k} 
        \left(\prod_{j=i}^k {\frac{\partial \x_j}{\partial \x_{j-1}}}\right)
        \frac{\partial \x_i}{\partial {\d}}
    \right),
\end{equation}
where:
\begin{align}
    \frac{\partial \x_j}{\partial \x_{j-1}}  &= \left.\frac{\partial \bm{f}^{\prime}}{\partial \bm{x}} \right|_{(\bm{x}_{j-1},\bm{d})} = \left.( \frac{\partial\bm{f}}{\partial\bm{x}} + \frac{\partial\bm{f}}{\partial\bm{u}}\frac{\partial\bm{\pi}^*}{\partial\bm{x}} )\right|_{(\bm{x}_{j-1},\bm{d})}\label{eq:gradient_x} \\ 
    \frac{\partial \x_i}{\partial \bm{d}}  &= \left.\frac{\partial \bm{f}^{\prime}}{\partial \bm{d}} \right|_{(\bm{x}_{i-1},\bm{d})} = \left.( \frac{\partial\bm{f}}{\partial\bm{d}} + \frac{\partial\bm{f}}{\partial\bm{u}}\frac{\partial\bm{\pi}^*}{\partial\bm{d}} )\right|_{(\bm{x}_{i-1},\bm{d})} \label{eq:gradient_d}.
\end{align}

Here, $\frac{\partial \bm{f}^{\prime}}{\partial \bm{x}} \in \mathbb{R}^{n\times n}$ represents the closed-loop dynamics Jacobian while the $\frac{\partial \bm{f}}{\partial \bm{x}} \in \mathbb{R}^{n\times n}$ denotes the open-loop dynamics Jacobian, both with respect to the previous state. $\frac{\partial\bm{f}}{\partial\bm{u}} \in \mathbb{R}^{n\times m}$ denotes the open-loop dynamics Jacobian with respect to the control vector while $\frac{\partial\bm{\pi}^*}{\partial\bm{x}} \in \mathbb{R}^{m\times n}$ represents the Jacobian of the control policy with respect to the current state. The Jacobian with respect to the design parameters $\bm{d}$ follows a similar definition.
To compute the gradient above, the dynamics $\bm{f}$, the design-aware control policy $\bm{\pi}^*$, and cost function $l_k$ need to be differentiable with respect to the system state $\bm{x}$, the control input $\bm{u}$, and the design parameters $\bm{d}$. Under this regularity condition, first-order optimization methods leveraging gradient information can be employed, leading to faster convergence and improved reproducibility.

\section{SurGE Co-design for Legged Robots}

This section presents SurGE, a differentiable co-design framework (Fig.~\ref{fig:differentiable_pipeline}) that leverages a differentiable surrogate dynamics model to obtain the gradient information despite the discontinuities of legged robots' dynamics. Combined with a differentiable design-aware control policy, we derive a surrogate gradient of the design objective with respect to the design parameters, which is incorporated into the CMA-ES method to achieve fast and robust convergence. The development of the control policy, the derivation of the surrogate gradient, and the evolutionary search algorithm are detailed in the following subsections.

\subsection{Design-Aware Control Policy}\label{sec:design_aware_policy}
\begin{figure}[htb]
    \centering
    \includegraphics[width=1.0\linewidth]{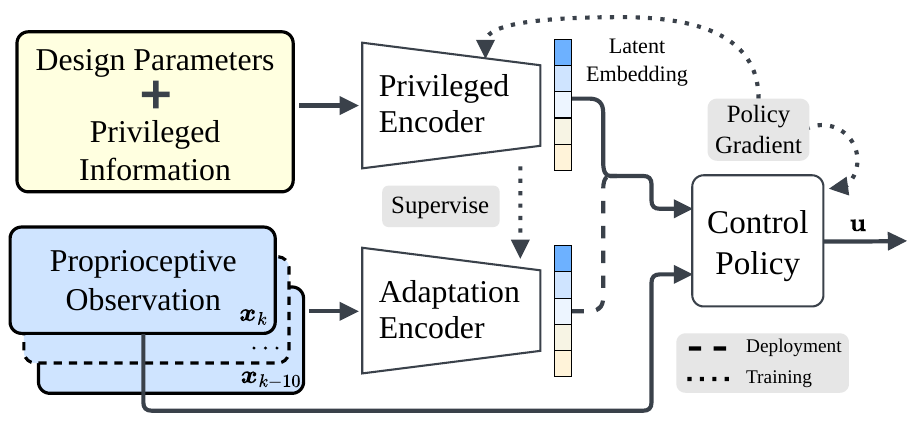}
    \caption{The design-aware control policy architecture. The design parameters are informed via a privileged encoder for training and design optimization, while the adaptation encoder is used for deployment. All neural networks are trained simultaneously thanks to ROA.}
    \label{fig:rl_policy}
    \vspace{-0.4cm}
\end{figure}

A differentiable co-design framework requires the design-aware control policy to be differentiable with respect to the system state and design parameters according to the gradient equations in Eq.\eqref{eq:gradient_x} and \eqref{eq:gradient_d}. 
Here, we use an RL policy as the controller since the neural networks are inherently differentiable.
To enable the policy to be design-aware, the normalized design parameters are encoded as privileged information~\cite{lee2020learning} via a multi-layer perceptron (MLP) and concatenated with the observation to form the policy input.
A teacher-student architecture with regularized online adaptation (ROA)~\cite{fu2023deep} is adopted for sim-to-real transfer.
The architecture of the policy is shown in Fig.~\ref{fig:rl_policy}.
Then, the policy is pretrained with Proximal Policy Optimization (PPO)~\cite{schulman2017proximal} across uniformly sampled designs per episode, encouraging generalization over the whole design space $\DS$.
By generalizing across $\DS$, this single policy aims to approximate the per-design optimal controller $\bm{\pi}^*$ across the design space, amortizing the inner loop and reducing the need for per-design retraining during the outer search~\cite{bjelonic2023learning}.

\subsection{Surrogate Gradient}\label{sec:gradient_computation}

Legged robot are subject to nonlinear hybrid dynamics, which is difficult to differentiate to obtain the gradient. Nevertheless, the gradient of a surrogate model that captures the dominant dynamical effect of the robot still provides useful information while preserving differentiability.
For example, the Single Rigid Body (SRB) model is often employed as the surrogate dynamics model for quadrupedal robots \cite{ding2019real}. Song et al. \cite{song2024learning} demonstrated that an effective surrogate model assists gradient-based optimization. 

Notably, the surrogate model state $\bm{s}$ is often either a subset or a function of the full state $\bm{x}$. In the case where $\bm{s}\subset\bm{x}$, the surrogate model ensures that only $\bm{s}$ is propagated through the differentiable dynamics for the gradient computation. The remaining states $\bm{x}\setminus\bm{s}$ are computed via the non-differentiable simulator and treated as constants with respect to $\d$.
Since the surrogate model only captures the dominant dynamics, its gradient is a biased approximation of the true gradient. Nevertheless, CMA-ES is robust to biased gradients, as the unbiased objective is evaluated by the non-differentiable simulator, which grounds the evolutionary process.

\subsection{State Alignment}\label{sec:state_alignment}
Due to the simplification of the surrogate model, its dynamics integration result can diverge from the true robot dynamics over long horizons, causing inaccurate design objectives and gradient estimates.
Inspired by~\cite{song2024learning}, the differentiable state $\x^{\text{diff}}$ is realigned at every time step with the state $\x^{\text{nd}}$ produced by a higher-fidelity non-differentiable simulator.
During the forward pass, the surrogate state $\bm{s}$ is extracted from $\x^{\text{nd}}$ so that each one-step prediction starts from a realistic state.
During the backward pass, the gradient is propagated through each one-step prediction.
This limits gradient accumulation to single-step horizons, preventing drift from model mismatch while still providing informative directional signals.

\subsection{Gradient-Guided Evolutionary Search}\label{sec:gradient_injection}

CMA-ES maintains a multivariate normal distribution $\mathcal{N}(\bm{m}_g, \sigma_g^2 \C_g)$ over the design space, where $\bm{m}_g$ is the mean, $\sigma_g$ the step size, and $\C_g$ the covariance matrix at generation $g$.
At each generation, $\lambda$ candidate solutions are sampled from this distribution, evaluated on the true objective $\LS$, and the distribution parameters are updated based on ranked fitness values.

To incorporate the surrogate gradient $\nabla_{\d}\LS$ computed by the differentiable pipeline (Sec.~\ref{sec:gradient_computation}), injection via mean shift~\cite{hansen2011injecting} is applied: the CMA-ES mean is shifted in the negative gradient direction before sampling.
The shift is preconditioned by the current covariance matrix and scaled to match the expected Mahalanobis step length of a regular CMA-ES sample:
\begin{equation}\label{eq:delta_mean}
    \Delta \bm{m}_g = - \frac{\sigma_g \sqrt{n} \C_g\nabla_{\d}\LS}{\|\C_g^{\half}\nabla_{\d}\LS\|},
\end{equation}
where the preconditioned direction $\C_g\nabla_{\d}\LS$ acts as a natural gradient step that accounts for the local geometry of the search distribution, while the $\sqrt{n}$ factor normalizes the Mahalanobis norm to the expected length $\chi_n$ of an $n$-dimensional standard normal vector.

Since the surrogate gradient is an approximation of the true gradient, it may accumulate bias and interfere with the CMA-ES adaptation mechanism. To mitigate this effect, we use a cosine-annealed injection rate that scales the mean shift
\begin{equation}\label{eq:decay}
    \eta_g = \tfrac{\eta_0}{2}\bigl(1 + \cos(\pi g / G)\bigr),
\end{equation}
where $\eta_0$ is the initial gradient step size and $G$ is the total number of generations.
This schedule emphasizes gradient guidance in early generations, when the surrogate provides the most benefit by steering the search toward promising regions, and gradually switches control to the CMA-ES adaptation as the distribution contracts around a local optimum.
The importance of this decay is validated experimentally in Sec.~\ref{sec:opt_results}.

The SurGE co-design framework is shown in Algorithm~\ref{alg:codesign_algo}.
At each generation, the surrogate gradient is computed at the current CMA-ES mean and used to shift the mean.
The $\lambda$ candidates are then sampled from the shifted distribution, evaluated in the non-differentiable simulator, and CMA-ES is updated with the fitness values.

\begin{algorithm}[htb]
    \caption{Surrogate Gradient Guided Co-design}
    \label{alg:codesign_algo}
    \begin{algorithmic}[1]
    \Require Initial design $\d_0$, design iterations $G$, population size $\lambda$, pretrained policy $\pi(\cdot, \d)$ 
    \Ensure Optimized design $\d^*$
    \State Initialize CMA-ES with $\bm{m}_0 \gets \d_0$\
    \State $\d^* \gets \d_0$, \ \ $\LS^* \gets \infty$
    \For{$g = 0 \textbf{ to } G-1$}
        \State $\nabla_{\d}\LS \gets \textsc{ComputeGradient}(\bm{m}_g, \pi)$ \Comment{Sec.~\ref{sec:gradient_computation}}
        \State $\bm{m}_g \gets \bm{m}_g + \eta_g \Delta \bm{m}_g$ \label{alg:meanshift_injection} \Comment{Eq.~\eqref{eq:delta_mean}}
        \State $\{\d_1, \ldots, \d_\lambda\} \gets \text{CMA-ES.sample}(\bm{m}_g)$
        \State $\LS_p \gets \textsc{Evaluate}(\d_p, \pi)$ for $p = 1,\ldots,\lambda$ \Comment{Eq.~\eqref{eq:design_objective_high_level}}
        \State $\bm{m}_{g+1} \gets \text{CMA-ES.update}(\{\d_p, \LS_p\}_{p=1}^{\lambda})$
        \State $p^\star \gets \argmin_{p} \LS_p$
        \If{$\LS_{p^\star} < \LS^*$}
            \State $\LS^* \gets \LS_{p^\star}$, \ \ $\d^* \gets \d_{p^\star}$
        \EndIf
    \EndFor
    \State \Return $\d^*$
    \end{algorithmic}
\end{algorithm}

\section{SurGE Co-design for MUPS}
\begin{figure}[htb]
    \centering
    \includegraphics[width=0.9\columnwidth]{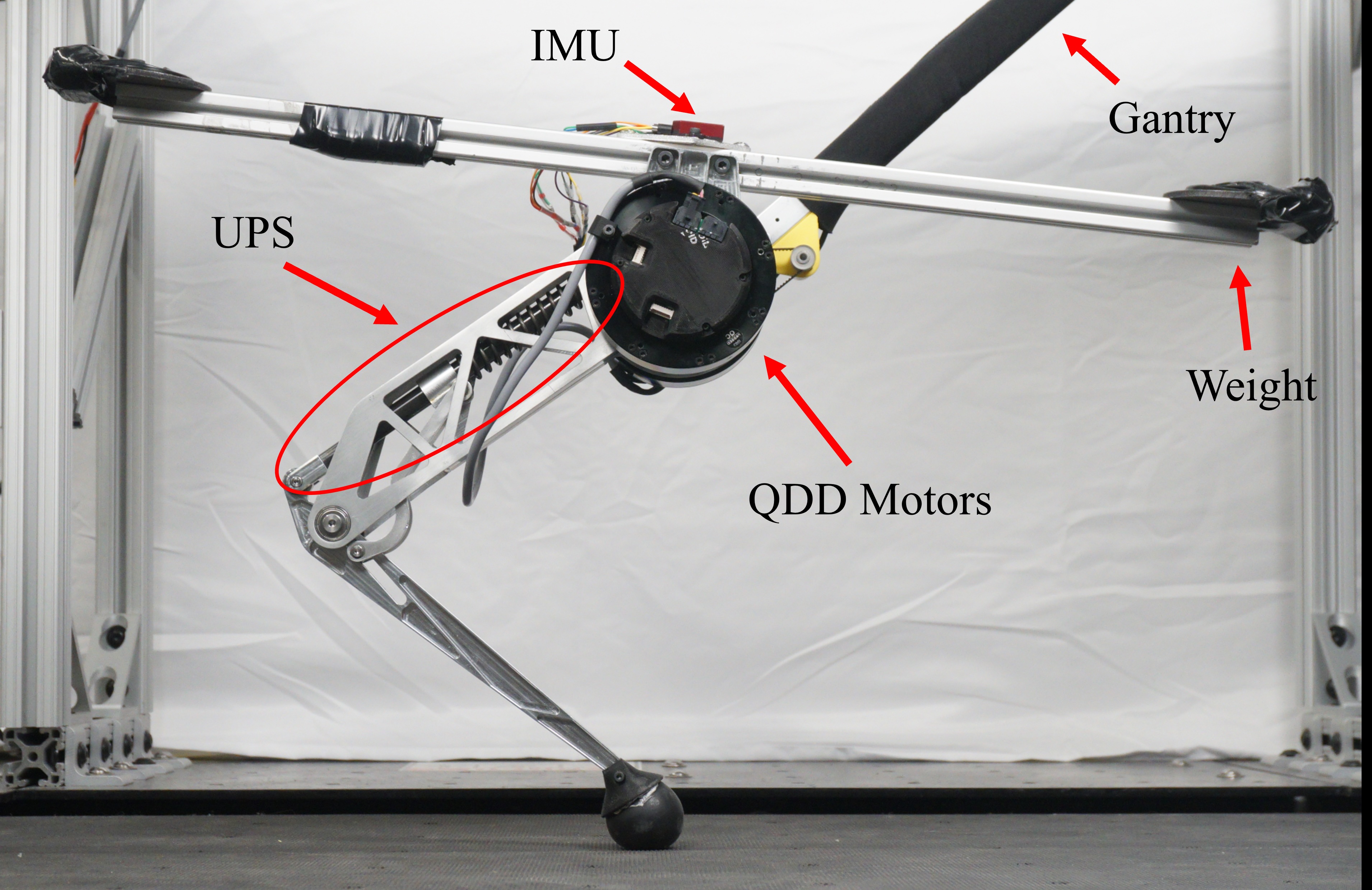}
    \caption{The Monoped with Unidirectional Parallel Spring v2 (MUPS~v2) is a customized hopping robot used for validating the SurGE framework.}
    \label{fig:mups_v2}
    \vspace{-0.5cm}
\end{figure}

The MUPS~v2, shown in Fig.~\ref{fig:mups_v2}, is a planar monoped hopping robot with a hip and a knee motor and a free-rotating torso constrained to the sagittal plane by a gantry.
Its UPS mechanism allows continuous adjustment of all four design parameters (Table~\ref{tab:design_param}), enabling hardware validation of the optimized designs.

\subsection{UPS Design}

The UPS on MUPS~v2 consists of a linear compression spring connected to the knee joint through a parallel linkage mechanism, as shown in Fig.~\ref{fig:cover_img_merged}~(b).
A typical torque-angle curve is shown in Fig.~\ref{fig:cover_img_merged}~(c), whose shape is determined by four design parameters, namely the spring stiffness $k_s$, the spring engagement length $l_0$, the rocker length $l_{r}$, and the crank link, which is the offset of the rocker attachment point on the shin from the knee joint $l_{c}$. The engagement point is smoothed via a \texttt{softplus} function.
These parameters fully characterize the engagement threshold, peak torque, and torque curve shape of the UPS, while the remaining mechanical dimensions are fixed by the existing hardware.
All four parameters are continuous and bounded within the ranges listed in Table~\ref{tab:design_param}, which are determined by the physical constraints of the robot.

\begin{table}[htb]
    \centering
    \caption{Design Parameters of MUPS~v2 Robot}
    \label{tab:design_param}
    \begin{tabular}{lccc}
    \toprule
    Parameter & Symbol & Range & Unit\\ 
    \midrule
    Spring Stiffness & $k_s$ & $[1000, 8000]$ & N/m\\
    Spring Engagement Length & $l_0$ & $[0.1, 0.17]$ & m\\
    Rocker Length & $l_{r}$ & $[0.08, 0.14]$ & m\\
    Crank Length & $l_{c}$ & $[0.02, 0.04]$ & m\\
    \bottomrule
    \end{tabular}
\end{table}

\subsection{Design Objective}
The design objective is formulated to capture both task performance and energy efficiency. Specifically, we evaluate the accumulated tracking error and heat dissipation over a finite trajectory. The robot is commanded to hop at a fixed apex height while tracking a fixed target speed on flat ground.
The per-step cost $l$ consists of a heating energy term and a tracking penalty:
\begin{equation}\label{eq:design_objective}
    l_k = \ln(1 + P_{\text{heat}, k} \Delta t) + \beta\|\x_k^{\des} - \x_k\|^2,
\end{equation}
where 
$P_{\text{heat}} = \|\btau_m\|^2 R / K_t^2$ is the power dissipated by Joule heating;
$\btau_m\in\real^2$ is the motor torque;
$\ln(1+x)$ is used to preserve stable gradients for small values of $x$;
$R$ and $K_t$ are the motor resistance and torque constant;
and $\beta$ weights the tracking term.
The desired state $\x_k^{\des}$ specifies the target base height and forward velocity, which are held constant across time steps.
The total objective $\LS(\d) = \sum_{k=1}^{N} l_k$ is accumulated over $N$ simulation steps.

\subsection{Kinodynamic SRB with Smoothed Spring Function}
Under our scenario, a kinodynamic single-rigid-body (Kino-SRB) model~\cite{zhuang2025kinodynamic} is adopted as our differentiable surrogate.
The Kino-SRB captures the essential centroidal dynamics of MUPS~v2 and incorporates the UPS torque as a smooth function of the design parameters, providing a compact differentiable path from $\d$ to $\LS$.
The SRB state $\x = [\p_c, \theta, \dot{\p}_c, \dot{\theta}]\T$ evolves as
\begin{equation}\label{eq:srb}
\dot{\bm{x}}
=
\begin{bmatrix}
    \dot{\p}_c \\
    \dot{{\theta}} \\
    \mathbf{f}/m+\bm{g} \\
    (\bm{r} \wedge \mathbf{f})/{I}
\end{bmatrix},
\end{equation}
where $\p_c\in\real^2$ and $\theta\in\real$ are the center of mass (CoM) position and body pitch;
$\mathbf{f}\in\real^2$ is the ground reaction force (GRF) at the foot;
$m$ is the total mass;
$\bm{g}$ is the gravity vector;
$\bm{r}=\p_f - \p_c$ is the CoM-to-foot vector;
and $I$ is the body moment of inertia.
A proportional–derivative (PD) controller is used to convert desired joint positions from policy output to motor torques.
Under a massless-leg assumption, the joint torque combines motor and spring contributions and maps to the GRF through the leg Jacobian:
\begin{equation}\label{eq:srb_torque}
    \btau = \btau_m + \btau_s(\bm{q}, \d) = -\B \cdot \J\T \f,
\end{equation}
where $\btau_m\in\real^2$ is the motor torque;
$\btau_s$ is the spring torque modeled as a differentiable function of joint positions $\bm{q}\in\real^2$ and design parameters $\d$;
$\B\in\real^{2\times 5}$ is the actuated-joint selection matrix;
and $\J\in\real^{2\times 5}$ is the leg Jacobian.

\section{Experiment Results}
\subsection{Computation Setup}\label{sec:experiment_setup}
The design-aware control policy is pretrained in IsaacGym~\cite{makoviychuk2021isaac} with GPU-accelerated parallel rollouts.
The policy is trained to hop on flat ground at fixed target speed and height, matching the conditions used for design evaluation.
The same simulator serves as the non-differentiable evaluator during the co-design optimization, computing the true design objective $\LS$ for each candidate.
The hyperparameters used for training are shown in Tab.~\ref{tab:rl_hyperparam} and reward terms are shown in Tab.~\ref{tab:rl_reward}, where $c$ is the contact indicator.

SurGE is compared against three baselines: (1) surrogate gradient descent (GD), which directly minimizes the surrogate objective via Adam; (2) vanilla CMA-ES; and (3) SurGE without cosine decay (SurGE w/o decay), which applies a constant injection step $\eta_0$.
Each stochastic method is run for 5 random seeds, while GD is deterministic.
The optimization hyperparameters are listed in Table~\ref{tab:hyperparams}.
Performance is measured by three metrics: the best-so-far objective (BSF), which tracks the minimum objective function value discovered over optimization iterations, the design objective gap $\bar{\LS} - \LS^*$ between the population mean and best at the final generation, and the per-iteration wall-clock time.

\begin{table}[htb]
    \centering
    \caption{Optimization Hyperparameter}
    \label{tab:hyperparams}
    \begin{tabular}{lcc}
    \toprule
    Hyperparameter              & Symbol     & Value \\
    \midrule
    Tracking weight             & $\beta$    & 5 \\
    Design iterations           & $G$        & 50  \\
    Control steps               & $N$        & 100 \\
    Population size             & $\lambda$  & 16  \\
    Initial search step    & $\sigma_0$ & 0.2 \\
    Initial gradient step  & $\eta_0$   & 0.1 \\
    \bottomrule
    \end{tabular}
    \vspace{-0.3cm}
\end{table}

\begin{table}[htb]
    \centering
    \caption{RL Training Hyperparameter}
    \begin{tabular}{lc}
        \toprule
        Hyperparameter & Value \\
        \midrule
        \multicolumn{2}{c}{\textbf{Network architecture}} \\
        Actor MLP hidden dimensions & $[128, 64, 32]$\\
        Critic MLP hidden dimensions & $[128, 64, 32]$\\
        Encoder MLP hidden dimensions & $[64, 20]$\\
        Latent embedding dimensions & $10$\\
        Activation function & $\texttt{ELU}$\\
        \multicolumn{2}{c}{\textbf{PPO}} \\
        Learning rate & $2\times10^{-4}$\\
        Discount factor & $0.99$\\
        GAE discount factor & $0.95$\\
        Desired KL-divergence & $0.01$\\
        Clip range & $0.2$\\
        Entropy coefficient & $0.01$\\
        Value function loss coefficient & $1.0$\\
        Steps per environment & $24$\\
        Number of environments & $4096$\\
        Number of epochs per update & $5$\\
        \bottomrule
    \end{tabular}
    \label{tab:rl_hyperparam}
    \vspace{-0.4cm}
\end{table}

\begin{table}[htb]
    \centering
    \caption{RL Reward Terms Summary}
    \begin{tabular}{lcc}
        \toprule
        Reward Term & Definition & Weight\\
        \midrule
        \multicolumn{3}{c}{\textbf{Tracking reward}} \\
        Base height & $\exp(-\|\bm{p}_z^{\mathrm{err}}\|/0.25^2)$ & $2.0$\\
        Base orientation & $\exp(-\|\theta^{\mathrm{err}}\|/0.25^2)$ & $1.0$\\
        Base linear velocity & $\exp(-\|\bm{v}^{\mathrm{err}}\|/0.25^2)$ & $1.0$\\
        Base angular velocity & $\exp(-\|\bm{\omega}^{\mathrm{err}}\|/0.25^2)$ & $0.5$\\
        \multicolumn{3}{c}{\textbf{Regularization penalty}} \\
        Action smoothness & $-\|\bm{u}_k - \bm{u}_{k-1}\|^2$ & $0.1$\\
        Torque smoothness & $-\|\bm{\tau}_{m, k} - \bm{\tau}_{m, k-1}\|^2$ & $1\times 10^{-7}$\\
        Joint torque & $-\|\bm{\tau}_m\|^2$ & $5\times 10^{-3}$\\
        Joint acceleration & $-\|\ddot{\bm{q}}\|^2$ & $5\times 10^{-5}$\\
        Joint position & $-\|{\bm{q} - \bm{q}^{\mathrm{default}}}\|^2$ & $0.02$\\
        Undesired contacts & $-\mathbf{1}_{\{\mathbf{f} > 0.1\}}$& $10$\\
        Jumping upward & 
            $[\bm{v}_z]_{+}^2 \mathbf{1}_{\{\bm{p}_z < \bm{p}_z^{\mathrm{des}}\}}$
            & $0.1$\\
        Phase contact & 
            $\begin{cases}
                +1, & c=c^{\mathrm{des}}\\
                -1, & c\neq c^{\mathrm{des}}
            \end{cases}$ 
            & $1.0$\\
        \bottomrule
    \end{tabular}
    \label{tab:rl_reward}
    \vspace{-0.4cm}
\end{table}

\subsection{Surrogate Gradient Validation}
To validate the quality of the surrogate gradient before using it in SurGE, we visualize the gradient field over the design objective landscape.
Fig.~\ref{fig:cover_img_merged} shows a 2D slice through the $(k_s, l_0)$ subspace, with the remaining parameters fixed at the center of their ranges.
The contour surface is obtained by evaluating the non-differentiable simulator on a $32 \times 32$ grid, while the arrows are the surrogate gradients computed via the differentiable Kino-SRB pipeline (Sec.~\ref{sec:gradient_computation}).
Despite the model mismatch between the SRB surrogate and the full-body simulator, the surrogate gradients point predominantly downhill toward the low-objective region.
This result confirms that the surrogate pipeline provides a useful and directional prior.

\subsection{SurGE Design Optimization}\label{sec:opt_results}

The SurGE framework is first applied on a simple 2D design space ($k_s$, $l_0$) with the remaining parameters fixed.
Fig.~\ref{fig:cover_img_merged} overlays the trajectories of SurGE and CMA-ES on the true objective landscape.
The CMA-ES trajectory follows a zigzag path with frequent directional changes, reflecting its stochastic nature, whereas the SurGE trajectory is steered toward the low-objective region from the first few generations, converging with fewer wasted evaluations.

\begin{figure}[htb]
    \centering
    \includegraphics[width=1.0\columnwidth]{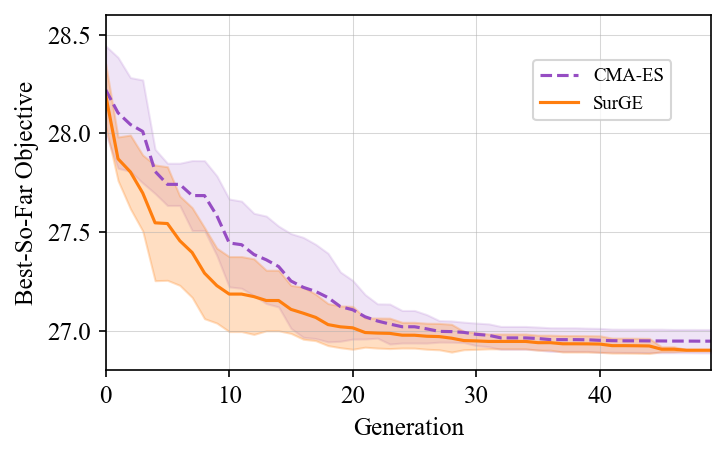}
    \caption{Comparison of BSF objective across generations. The shaded area represents the standard deviation across 5 random seeds.}
    \label{fig:batch_analysis_guided_es}
    \vspace{-0.4cm}
\end{figure}

Table~\ref{tab:design_opt_results} summarizes the full 4D optimization results.
SurGE achieves a mean BSF objective of $26.90$, comparable to vanilla CMA-ES while reducing cross-seed standard deviation by $6\times$.
The BSF curves in Fig.~\ref{fig:batch_analysis_guided_es} show that SurGE attains lower objective values in earlier generations, indicating that the surrogate gradient accelerates convergence throughout the search.
The design objective gap $\bar{\LS} - \LS^*$ also drops from $65.9$ (CMA-ES) to $53.9$ (SurGE), an $18\%$ reduction, indicating that the gradient-informed mean shift concentrates the population around higher-performing designs. 
In terms of computation, SurGE adds one surrogate-gradient rollout and backpropagation per generation. This increases the per-iteration time from $0.58$~s for CMA-ES to $1.27$~s for SurGE (Table~\ref{tab:design_opt_results}), while the full optimization still completes in about one minute. This roughly $2\times$ per-iteration overhead is a favorable trade-off for the improved convergence and reproducibility that SurGE provides.

\begin{table}[htb]
    \centering
    \caption{Design Optimization Results in 4D Design Space (5 seeds)}
    \label{tab:design_opt_results}
    \begin{tabular}{lccc}
    \toprule
    Method          & BSF $\downarrow$          & $\bar{\LS} - \LS^*$ $\downarrow$ & Time\,/\,Iter. [s] $\downarrow$ \\
    \midrule
    GD              & $27.08$                   & ---             & $1.27$ \\
    CMA-ES          & $26.95 \pm 0.06$          & $65.9$          & $\mathbf{0.58}$ \\
    SurGE w/o decay    & $26.93 \pm 0.03$          & $66.5$          & $1.27$ \\
    SurGE (ours)       & $\mathbf{26.90 \pm 0.01}$ & $\mathbf{53.9}$ & $1.27$ \\
    \bottomrule
    \end{tabular}
\end{table}

Ablation studies were conducted to isolate the contribution of cosine decay and evolution strategy, with results included in Table~\ref{tab:design_opt_results}. Removing cosine decay (SurGE w/o decay) degrades BSF and increases cross-seed standard deviation by $3\times$, while the population fitness gap rises from $53.9$ to $66.5$.
This confirms that annealing the injection step size is critical: without decay, the persistent gradient bias interferes with the CMA-ES adaptation in later generations.
Surrogate GD, which follows the surrogate gradient exclusively via Adam, yields the worst objective ($27.08$) and converges to a qualitatively different local minimum, demonstrating that population-based exploration is essential for navigating the non-convex design landscape.

\begin{figure}[htb]
    \centering
    \includegraphics[width=1.0\columnwidth]{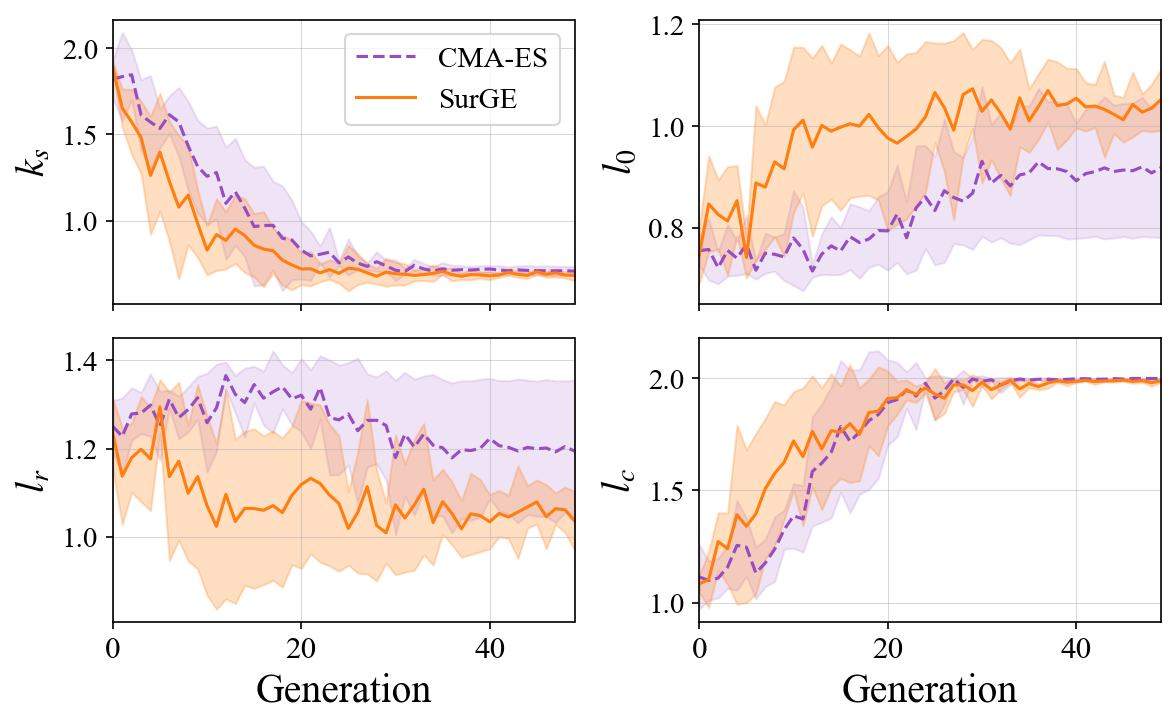}
    \caption{Design parameter trajectories across generations. The shaded area represents the standard deviation across 5 random seeds. Values are shown in scaled units.}
    \label{fig:design_params_cma_vs_ms01d100}
\end{figure}

As shown in Fig.~\ref{fig:design_params_cma_vs_ms01d100}, the improved reproducibility is most pronounced on the spring-length parameters where the guidance signal of surrogate gradient is strongest. The cross-seed standard deviation of $l_0$ and $l_{r}$ drops by $2.3\times$ compared to CMA-ES, indicating that SurGE steers the search toward a consistent basin, reducing sensitivity to random initialization.

\subsection{Hardware Experiments}
\begin{figure}[htb]
    \centering
    \includegraphics[width=0.8\linewidth]{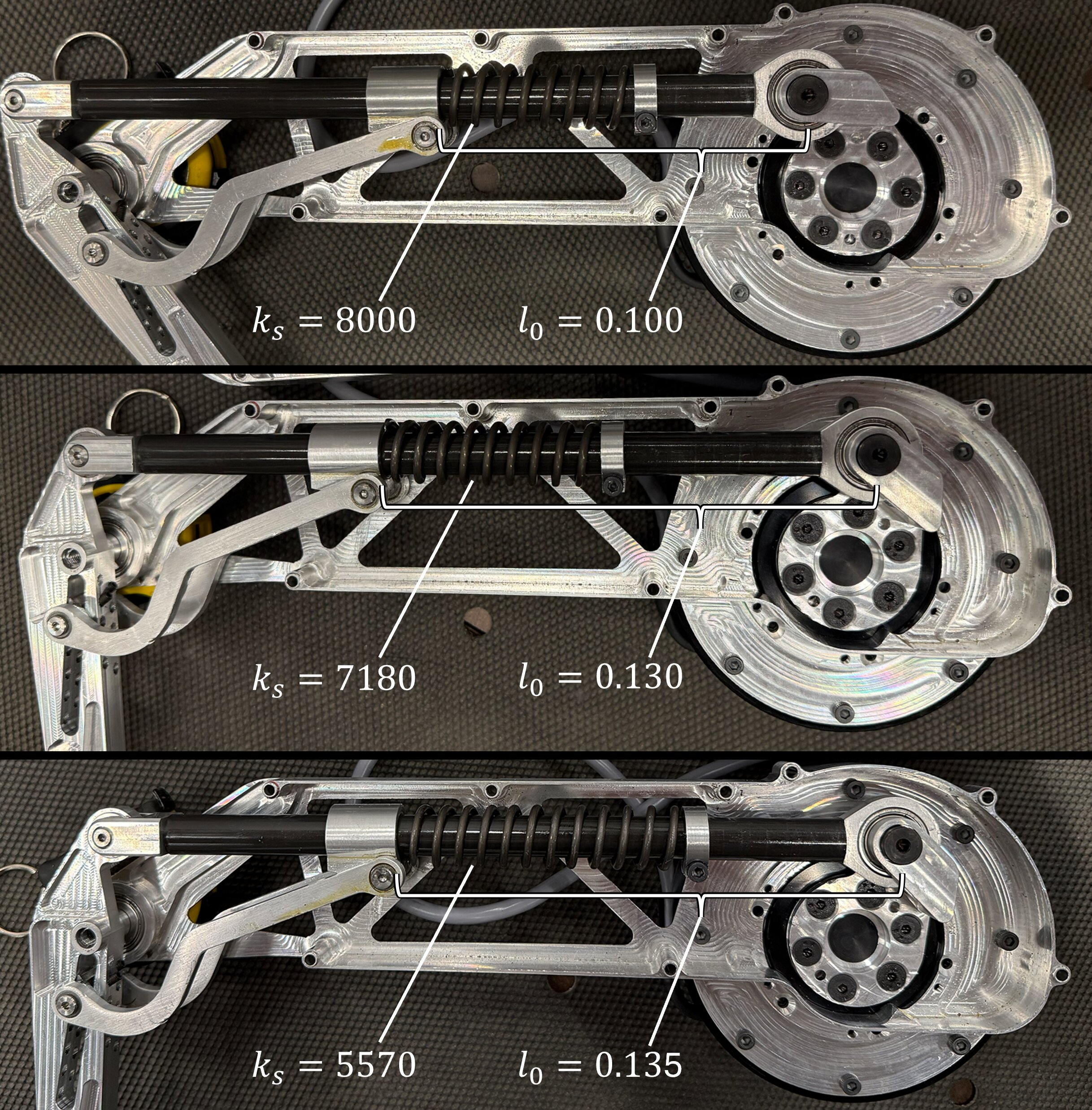}
    \caption{MUPS v2 hopper assemblies for the 3 designs of UPS tested for the hardware experiments. $k_s$ is spring stiffness and $l_0$ is spring rest length.}
    \label{fig:three_point_sweep_replacements_ps_text}
\end{figure}

Hardware experiments were conducted on the MUPS-v2 robot hardware to validate the optimization results. Specifically, 3 designs along the SurGE optimization path were selected, including the hand-tuned initial design, an intermediate design (gen 5), and the final optimized design.
For each configuration, the spring parameters are physically adjusted on the robot as shown in Fig.~\ref{fig:three_point_sweep_replacements_ps_text} and the controller is deployed to hop at the same target speed used in simulation.

The trained policy is deployed on a workstation with a 13th Gen Intel Core i7-13700 CPU, which sends torque commands to the robot via LCM at 50~Hz.
On board, a Raspberry Pi relays commands to the motors, reads motor encoders and an IMU, and runs a Kalman filter at 1~kHz to estimate the robot state.


\begin{table}[htb]
    \centering
    \caption{3-Point Sweep Hardware Validation}
    \label{tab:hardware_validation_2d}
    \begin{tabular}{lcccc}
    \toprule
    Gen. & $k_s$ [N/m] & $l_0$ [m] & $\bar{\LS}_{\text{sim}}$ & $\bar{\LS}_{\text{hw}}$ \\
    \midrule
    $g=1$   & $8000$ & $0.100$ & $38.42$  & $44.68$ \\
    $g=5$   & $7180$ & $0.130$ & $28.04$  & $28.79$ \\
    $g=20$  & $5570$ & $0.135$ & $27.64$  & $27.86$ \\
    \bottomrule
    \end{tabular}
\end{table}

Table~\ref{tab:hardware_validation_2d} reports the design objectives evaluated in both simulation and on hardware for three design points sampled from the 2D optimization trajectory in Fig.~\ref{fig:cover_img_merged}.
The hardware objective decreases monotonically along the trajectory, matching the trend in simulation, confirming that the optimization direction identified by SurGE transfers to the physical system.
Overall, starting from the hand-tuned initial design, SurGE reduces the design objective by $28.06\%$ in simulation, while achieving a $37.65\%$ reduction on hardware. 
The hand-tuned design serves only as an initialization that SurGE refines. The larger hardware improvement is attributable to the elevated sim-to-real gap at the initial design point.

\section{Conclusions}\label{sec:conclusion}

This work presents SurGE, a surrogate gradient-guided evolution co-design framework for legged robots with parallel elasticity. Optimization is performed on a custom monoped hopping robot named MUPS-v2. Our key contribution is the proposition of using the surrogate gradients through a differentiable pipeline, which consists of a kinodynamic SRB model and a design-aware control policy, to shift the CMA-ES mean towards minima. To mitigate the surrogate bias, a cosine decay gradually cedes control to the CMA-ES adaptation as the optimization converges. Compared to vanilla CMA-ES, SurGE matches the best objective and achieves $6\times$ lower cross-seed standard deviation and $18\%$ tighter population concentration in a 4D design space. Hardware validation on the MUPS~v2 robot demonstrates that the optimized design transfers to the physical system, with a monotonic improvement in design objective along the optimization path.

Two limitations motivate future work.
First, the surrogate gradient is biased by the model mismatch, and incorporating full-body dynamics with saltation matrices\cite{kong2024saltation} to differentiate through contact transitions could improve gradient fidelity.
Second, the control policy is pretrained and fixed during optimization, but refining the policy online is possible via model-based RL, which could directly leverage the surrogate gradient signal and further improve design quality.

\addtolength{\textheight}{-0cm}   





\bibliographystyle{IEEEtran}
\bibliography{reference}

\end{document}